%% file: main.tex
\documentclass{article}

\usepackage{microtype}
\usepackage{graphicx}
\usepackage{booktabs} 
\usepackage{hyperref}
\usepackage{xspace}
\usepackage{amsfonts, amsmath, amssymb}
\usepackage{subcaption}
\usepackage[fixed]{fontawesome5} 
\usepackage{arydshln} 
\usepackage{dsfont} 
\usepackage{bm} 
\usepackage{enumitem}
\usepackage{soul} 

\usepackage[accepted]{icml2021}
\usepackage{xurl}
\icmltitlerunning{Grounding Language to Entities and Dynamics for Generalization in Reinforcement Learning}

\newcommand{\revision}[1]{#1}
\newcommand{\camrdy}[1]{#1}

\newcommand{\taskname}{\textsc{Messenger}\xspace}
\newcommand{\modelname}{\textsc{Emma}\xspace}
\newcommand{\gid}{G-ID\xspace}
\newcommand{\bos}{Mean-BOS\xspace}
\newcommand{\oracle}{O-Map\xspace}
\newcommand{\bayes}{BAM\xspace}
\newcommand{\txtpi}{txt2$\pi$\xspace}

\newcommand{\idref}[1]{{\texttt{\small{#1}}}}
\newcommand{\roleref}[1]{{\textit{{#1}}}}
\newcommand{\strref}[1]{{`{#1}'}}

\newcommand{\faCircleO}{\faCircle[regular]}

\DeclareMathOperator{\softmax}{\sigma}
\DeclareMathOperator*{\argmax}{arg\,max}

\begin{document}

\twocolumn[
\icmltitle{Grounding Language to Entities and Dynamics \\for Generalization in Reinforcement Learning}


\begin{icmlauthorlist}
\icmlauthor{Austin W. Hanjie}{princeton}
\icmlauthor{Victor Zhong}{uw}
\icmlauthor{Karthik Narasimhan}{princeton}
\end{icmlauthorlist}

\icmlaffiliation{princeton}{Computer Science, Princeton University, USA}
\icmlaffiliation{uw}{Computer Science, University of Washington, USA}

\icmlcorrespondingauthor{Austin W. Hanjie}{hjwang@cs.princeton.edu}

\icmlkeywords{Machine Learning, ICML}

\vskip 0.3in
]


\printAffiliationsAndNotice{}

\input{sections/abstract}

\input{sections/introduction}

\input{sections/related}

\input{sections/framework}

\input{sections/task}

\input{sections/model}

\input{figures/train_curve}
\input{sections/setup}

\input{figures/train_table}
\input{sections/results}

\input{sections/conclusion}

\bibliography{karthik, anthology, others}
\bibliographystyle{icml2021}

\clearpage
\appendix
\input{sections/appendix}

\end{document}

%% file: sections/abstract.tex
\begin{abstract}
We investigate the use of natural language to drive the generalization of control policies and introduce the new multi-task environment \taskname with free-form text manuals describing the environment dynamics.
Unlike previous work, \taskname does not assume prior knowledge connecting text and state observations --- the control policy must simultaneously ground the game manual to entity symbols and dynamics in the environment.
We develop a new model, \modelname (Entity Mapper with Multi-modal Attention) which uses an entity-conditioned attention module that allows for selective focus over relevant descriptions in the manual for each entity in the environment.
\modelname is end-to-end differentiable and learns a latent grounding of entities and dynamics from text to observations using only environment rewards.
\modelname achieves successful zero-shot generalization to unseen games with new dynamics, obtaining a 40\% higher win rate compared to multiple baselines.
However, win rate on the hardest stage of \taskname remains low (10\%), demonstrating
the need for additional work in this direction.


\end{abstract}

%% file: sections/introduction.tex
\section{Introduction}
\label{sec:introduction}
\begin{figure}[t]
    \centering
    \includegraphics[width=\linewidth]{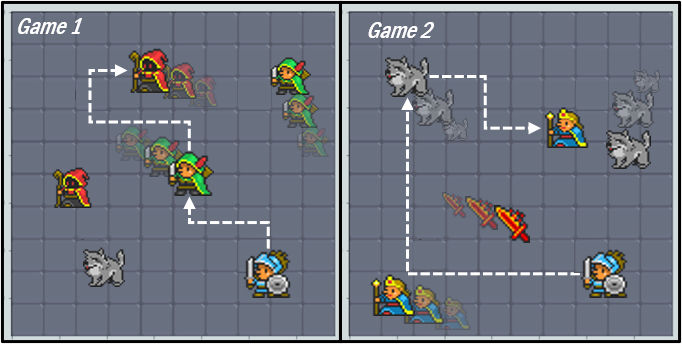}
    \begin{sc}
        \small
        Game 1 Manual
    \end{sc}
    \hrulefill
    \vspace{-0.5em}
    \begin{footnotesize}
    \begin{enumerate}[noitemsep, leftmargin=*]
        \item at a particular locale, there exists a motionless mongrel that is a formidable adversary.
        \item the top-secret paperwork is in the crook's possession, and he's heading closer and closer to where you are.
        \item the crucial target is held by the wizard and the wizard is fleeing from you.
        \item the mugger rushing away is the opposition posing a serious threat.
        \item the thing that is not able to move is the mage who possesses the enemy that is deadly.
        \item \textit{the vital goal is found with the canine, but it is running away from you.}
    \end{enumerate}
    \end{footnotesize}
    \vspace{-1em}
    \hrulefill
    \caption{Two games from our multi-task environment \taskname where the agent must obtain the message and deliver it to the goal (white dotted lines). Within a single game, the same entities (e.g.~mage) with different roles (e.g.~enemy, goal) must be disambiguated by their dynamics (e.g.~immovable, fleeing). The same entities may have different roles in different games forcing the agent to consult the manual to succeed consistently. Note the extraneous description (\textit{italics}) and multiple synonyms for entities and roles (e.g.~mage, wizard; adversary, opposition). Unlike prior work, the mapping from words in the manual to game entities is not available and must be learned using only scalar game rewards.
    }
    \label{fig:intro}
\end{figure}

Interactive game environments are useful for developing agents that learn grounded representations of language for autonomous decision making~\citep{golland-etal-2010-game,branavan2012learning,andreas2015alignment,bahdanau2018learning}.
The key objective in these environments is learning to interpret language specifications by relating entities and dynamics of the environment (i.e.~how entities behave) to their corresponding references in the text, in order to effectively and efficiently win new settings with previously unseen entities or dynamics~\citep{narasimhan2018grounding,zhong2020rtfm}.
While existing methods demonstrate successful transfer to new settings, they assume a ground-truth mapping between individual entities and their textual references.

We introduce \taskname,\footnote{Available at: \url{https://github.com/ahjwang/messenger-emma}} an environment which
features multiple game variants with differing dynamics and accompanying text manuals in English for each.
The manuals contain descriptions of the entities and world dynamics obtained through crowdsourced human writers. 
Crucially, while prior work assumes a ground truth mapping
(e.g.~the word \strref{knight} in the manual refers to the entity name \strref{knight} in the observation),
\taskname does not contain prior signals that map between text and state observations (e.g.~between the phrase \strref{mounted warrior is fleeing} and the symbol \faChessKnight~moving away from the agent).
To succeed in \taskname, an agent must relate entities and dynamics of the environment to their references in the natural language manual using only scalar reward signals from the environment.
The overall game mechanics of \taskname involve obtaining a \roleref{message} and delivering it to a \roleref{goal}. For instance, in game 1 of Figure~\ref{fig:intro}, the agent must read the manual to:
\begin{enumerate}[noitemsep, leftmargin=*]
    \item Identify the entity that holds the message. In this case, description 2 (d-2) reveals that it is with the \idref{thief} but there is an identical entity that is an enemy (d-4).
    \item Map d-2 and d-4 to the correct symbols in the observation (green-cloaked person).
    \item Observe the movement patterns of the two entities (\strref{heading closer} vs. \strref{rushing away}) to disambiguate which of the two entities holds the message.
    \item Pick up the message from the entity that holds it.
    \item Identify the entity that is the goal. Here, d-3 and d-6 reference a goal. It must realize that there is no \strref{canine} that is \strref{running away} and so d-6 must be a distractor, and a \idref{mage} must be the goal.
    \item Follow a similar procedure to 3 to disambiguate which \idref{mage} is the goal and which is the enemy (d-3 vs d-5).
    \item Bring the message to the goal.
\end{enumerate}



To ground entities and dynamics to their corresponding references in the manual, we develop a new model called \modelname (Entity Mapper with Multi-modal Attention).
\modelname simultaneously learns to select relevant sentences in the manual for each entity in the game as well as incorporate the corresponding text description into its control policy. 
This is done using a multi-modal attention mechanism which uses entity representations as queries to attend to specific tokens in the manual text. 
\modelname then generates a text-conditioned representation for each entity which is processed further by a deep neural network to generate a policy.
We train the entire model in a multi-task fashion using reinforcement learning to maximize task returns.

Our experiments demonstrate \modelname outperforms multiple baselines (language-agnostic, attention-ablated, and Bayesian attention) and an existing state of the art model~\citep{zhong2020rtfm} --- on unseen games (i.e. a zero-shot test), \modelname achieves more than $40\%$ higher win rates. However, while \modelname can effectively map text references to their corresponding entity symbols in observation space, its ability to disambiguate descriptions by grounding language to entity movement dynamics is lacking, and win rates on the test games for the hardest stage of \taskname remains low for all models evaluated ($\leq10\%$), demonstrating the challenging nature of grounding natural language to dynamics using only interactive (reward-based) feedback.

\camrdy{In summary, our paper makes two key contributions: (1) a multi-task environment with novel challenges including a) learning entity symbol grounding from scratch in a multi-task setup with b) realistic, crowd-sourced text and (2) an attention-based model that is able to learn such a grounding where prior approaches struggle.} We hope \taskname and \modelname will further enable the development of new models and learning algorithms for language grounding.



%% file: sections/related.tex
\section{Related Work}
\label{sec:related}

\paragraph{Grounding for Instruction Following}
Grounding natural language to policies has been explored in the context of instruction following in tasks like navigation~\citep{chen2011learning, hermann2017grounded, fried-etal-2018-unified,wang2019reinforced,daniele2017navigational,misra2017mapping,janner2018representation}, games~\citep{golland-etal-2010-game,reckman2010learning,andreas2015alignment,bahdanau2018learning,kuttler2020nethack} or robotic control~\citep{walter2013learning,hemachandra2014learning,blukis2019learning} (see~\citet{luketina2019survey} and \citet{tellex2020robots} for more detailed surveys). Recent work has explored several methods for enabling generalization in instruction following, including environmental variations~\citep{hill2019environmental}, memory structures~\citep{hill2020grounded} and pre-trained language models~\citep{hill2020human}. In a slightly different setting, \citet{co2018guiding} use incremental guidance, where the text input is provided online, conditioned on the agent’s progress in the environment. \citet{andreas2016modular} developed an agent that can use sub-goal specifications to deal with sparse rewards. \citet{oh2017zero} use sub-task instructions and hierarchical reinforcement learning to complete tasks with long action sequences.

 In all these works, the text conveys the goal to the agent (e.g.~\strref{move forward five steps}), thereby encouraging a direct connection between the instruction and the control policy. This tight coupling means that any grounding learned by the agent is likely to be tailored to the types of tasks seen in training, making generalization to a new distribution of dynamics or tasks challenging.
 In extreme cases, the agent may even function without acquiring an appropriate grounding between language and observations~\citep{hu2019you}.
 In our setup, we assume that the text only provides high-level guidance without directly describing the correct actions for every game state.

\paragraph{Language Grounding by Reading Manuals}
A different line of work has explored the use of language as an auxiliary source of knowledge through text manuals. These manuals provide useful descriptions of the entities in the world and their dynamics (e.g.~how they move or interact with other entities) that are optional for the agent to make use of and do not directly reveal the actions it has to take. \citet{branavan2012learning} developed an agent to play the game of Civilization more effectively by reading the game manual. They make use of dependency parses and predicate labeling to construct feature-based representations of the text, which are then used to construct the action-value function used by the agent. Our method does not require such feature construction. \citet{narasimhan2018grounding} and \citet{zhong2020rtfm} used text descriptions of game dynamics to learn policies that generalize to new environments, without requiring feature engineering. However, these works assume some form of initial grounding provided to the agent (e.g.~a mapping between entity symbols and their descriptions, or the use of entity names in text as state observations). In contrast, \taskname requires that this fundamental mapping between entity symbols in observation space and their text references be learned entirely through interaction with the environment.
%

%% file: sections/framework.tex
\section{Preliminaries}
Our objective is to demonstrate grounding of environment dynamics and entities for generalization to unseen environments. An \idref{entity} is an object represented as a symbol in the observation. Dynamics refer to how entities behave in the environment including how they interact with the agent. Notably, movement dynamics are the frame-to-frame position changes exhibited by entities (e.g. fleeing).

\paragraph{Environment}
We model decision making in each environment as a Partially-Observable Markov Decision Process (POMDP) with the 8-tuple $(S, A, O, P, R, E, Z, M)$. $S$ and $O$ are the set of all states and observations respectively where each $o\in O$ contains entities from the set of entities $E$. At each step $t$, the agent takes some action $a_t\in A$. $P( s_{t+1} | s_t, a_t)$ is the transition distribution over all possible next states $s_{t+1}$ conditioned on the current state $s_t$ and action $a_t$. $R(s_t, a_t, s_{t+1})$ is a function that provides the agent with a reward $r_t\in \mathbb{R}$ for action $a_t$ and transition from $s_t$ to $s_{t+1}$. $Z$ is a set of text descriptions, with each $z\in Z$ providing information about an entity $e \in E$. $M$ is the map $z_e \mapsto e$ which identifies the entity that each description describes. $M$, $P$, and $R$ are not available to the agent. Note that there might not be a one-to-one mapping between $Z$ and entities in the current state observation.

\paragraph{Reinforcement Learning (RL)}
The objective of the agent is to find a policy $\pi: O\rightarrow A$ to maximize its cumulative reward in an episode. If $\pi$ is parameterized by $\theta$,  standard deep RL approaches optimize $\theta$ to maximize the expected reward of following $\pi_\theta$. In our setup, we want the agent to learn a policy $\pi_\theta(a | o, Z)$ that conditions its behavior on the provided text. However, in contrast to previous work~\citep{narasimhan2018grounding,zhong2020rtfm}, $M$ is not available to our agent and must be learned.

\paragraph{Differentiating Entities, Roles, and Text References}
For ease of exposition, we use type face to differentiate between \idref{entity symbols}, \roleref{roles}, and \strref{text references}.
For example, \idref{plane}~refers to the entity \faPlane, where \strref{plane} and \strref{aircraft} are text references to \idref{plane}.
Additionally, \idref{plane} can take on the role of an \roleref{enemy}.

%% file: sections/task.tex
\section{\taskname}
We require an environment where grounding text descriptions $Z$ to dynamics and learning the mapping $M$ for all entities in $E$ is necessary to obtain a good reward. Moreover, there must be enough game instances of the environment to induce the mapping $M$.

With these requirements in mind, we devise a new multi-task environment \taskname using the Py-VGDL framework~\citep{schaul2013video}. In \taskname, each entity can take on one of three roles: an \roleref{enemy}, \roleref{message}, or \roleref{goal}. The agent's objective is to bring the message to the goal while avoiding the enemies. If the agent encounters an enemy at any point in the game, or the goal without first obtaining the message, it loses the game and obtains a reward of $-1$. Rewards of $0.5$ and $1$ are provided for obtaining and delivering the message to the goal respectively.\footnote{We find that our approach also works with sparser terminal $\pm1$ rewards (Fig. \ref{fig:terminal_rewards}, Appendix).}. There are twelve different entities and three possible movement types: stationary, chasing, or fleeing.
Each set of entity-role assignments (henceforth referred to as a game) is initialized on a $10\times 10$ grid. The agent can navigate via \idref{up}, \idref{down}, \idref{left}, \idref{right}, and \idref{stay} actions and interacts with another entity when both occupy the same cell.

The same set of entities with the same movements may be assigned different roles in different games. Thus, two games may have identical observations but differ in the reward function $R$ (which is not available to the agent) and the text manual $Z$ (which is available). Thus, our agent must learn to extract information from $Z$ to succeed consistently. Some game examples are presented in Figure \ref{fig:intro}.

\paragraph{Grounding Entities}
\taskname requires agents to learn $M$ without priors connecting state observations $O$ to descriptions $Z$. Aside from using independent entity symbols disjoint from the text vocabulary, the set of training games is designed such that simple co-occurrence statistics between entity and text do not completely reveal $M$.

Consider when every possible combination of entities is observed during training. Then, for an entity $e$, its symbol in the observation (e.g.~\idref{plane})  is the only one that always appears together with its text references (e.g.~\strref{aircraft}). This tight coupling provides an inherent bias towards the correct grounding without needing to act in the environment. We denote such a set of games where each entity can appear with every other entity as \textbf{multi-combination (MC)}.

The MC assumption may not always be realistic in practice --- some entities are very unlikely to appear together (e.g.~\idref{plane}, \idref{thief}, \idref{sword}) while others may co-occur exclusively with each other (e.g.~\idref{mage}, \idref{orb}, \idref{sword}). We denote games in which the same entities always appear together as \textbf{single-combination (SC)}.
For SC games, every text symbol in the manual (e.g.~\strref{mage}, \strref{enemy}, \strref{the}, etc.) co-occurs the same number of times with all entity symbols in the observation. \camrdy{For example, if the entity symbols} \faCircleO and \faSquare always appear simultaneously with both text symbols \strref{mage} and \strref{sword}, it is impossible to map \strref{mage} to \faCircleO without interacting with the entities. \camrdy{That is, co-occurrences between entity and text symbols provide no information about $M$ and \textbf{the agent must ground these entities entirely via interaction}. To learn $M$ for the entities in this example, the agent must interact with} \faCircleO and if it obtains the message from it, it must infer from the description \strref{The mage has the message} that \faCircleO must be a \strref{mage}.


\begin{figure}[t]
    \centering
    \includegraphics[width=0.85\linewidth]{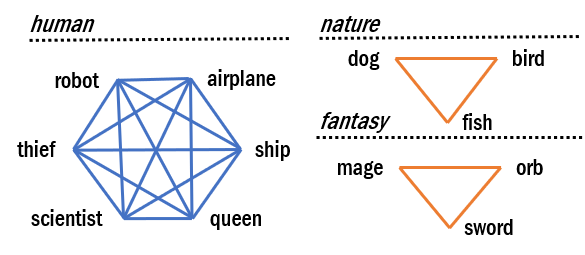}
    \caption{Entities and their subdivision into human, nature and fantasy sub-worlds. Each $K_3$ subgraph is a combination of entities that may appear during training. }\label{fig:entities}
\end{figure}

We divide the entities in \taskname into human, nature, and fantasy sub-worlds (Fig. \ref{fig:entities}) and exclude from training any games in which entities from different sub-world appear together. In particular, the nature and fantasy subworlds form SC and the human subworld forms the MC games.

\paragraph{Grounding Dynamics}
To force agents to distinguish varying movement dynamics, multiple copies of the same entity with different roles in \taskname may exhibit different movement patterns. For example, within the same game there may be descriptions: (1) \strref{the chasing mage is an enemy} and (2) \strref{the fleeing mage is the goal}. This means that even after grounding words such as \strref{mage} to its corresponding entity symbol, the agent must additionally consider the position of $e$ through a sequence of observations $o_{t-k},...,o_{t}$ in order to find the correct description $z_e$.

\paragraph{Text Descriptions}
We collected 5,316 unique free-form entity descriptions in English via Amazon Mechanical Turk \citep{buhrmester2016turk} by asking workers to paraphrase prompt sentences.
To increase the diversity of responses, the prompts were themselves produced from 82 crowdsourced templates. When constructing the prompts, we inject multiple synonyms for each entity. Workers  further paraphrased these synonyms, resulting in multiple ways to describe the same entity (e.g.~\strref{airplane}, \strref{jet}, \strref{flying machine}, \strref{aircraft}, \strref{airliner}). Furthermore, we observe responses with multiple sentences per description, typos (\strref{plane} vs \strref{plan}) and the need to disambiguate similar words (\strref{flying machine}, \strref{winged creature}).
Each training manual consists of a set of descriptions with an average total length of 30 - 60 words depending on the level. The total vocabulary size of the descriptions is 1,125. Besides lower-casing the worker responses, we do not do any preprocessing. Example descriptions can be found in Fig. \ref{fig:intro}. Further details regarding data collection can be found in appendix \ref{sec:text_details}.

\paragraph{Train-Evaluation Split} We ensure that any assignment of an entity to the roles \roleref{message} or \roleref{goal} in the evaluation games never appears during training (e.g.~if $e$ is the \roleref{goal} in evaluation, no $e$ is ever the \roleref{goal} in any training game). This forces models to make compositional entity-role generalizations to succeed on the evaluation games. In total we have 44 training, 32 validation, and 32 test games. We train on 2,863 of the text descriptions and reserve 1,227 and 1,226 for validation and testing respectively.

\paragraph{Comparison with Previous Environments}

\camrdy{We chose to realize} \taskname in a \camrdy{grid-world as it allows us to (1) study generalization to rich sets of procedurally generated dynamics, (2) conduct controlled studies of co-occurrence statistics (SC, vs. MC) and (3) explicitly verify the learned groundings with well-defined, discrete entities (see Fig.} \ref{fig:atten_heatmap}).

Other grid-worlds used to study language grounding include RTFM~\citep{zhong2020rtfm}, BabyAI~\citep{babyai_iclr19} and \citet{narasimhan2018grounding}. An oracle is used in \citet{narasimhan2018grounding} to concatenate the text representation to its corresponding entity representation. Access to such an oracle is a strong assumption in the wild and eliminates the need to ground the entities altogether.

In RTFM, the observation is a grid of text in which entity names are lexically identical to their references in the manual (e.g.~\strref{plane}).
The key challenge unique to \taskname is learning to map between the observed entity symbol (e.g.~\faPlane) and its natural language references in the manual (e.g.~\strref{aircraft}). 
Furthermore, RTFM is a \textbf{MC} environment
which may simplify the grounding problem.
Both \citet{narasimhan2018grounding} and \citet{zhong2020rtfm} do not consider disambiguation by grounding movement dynamics, whereas agents in Messenger need to distinguish entities based on how they move (e.g.~fleeing, chasing).

Unlike previous work on language grounding in grid environments \citep{zhong2020rtfm,babyai_iclr19}, we do not use templated or rule-generated text. RTFM uses a small number of rule-based templates to construct each manual, and each entity is referred to in a single way (e.g.~\idref{goblin} is always \strref{goblin}). In contrast, \taskname features thousands of hand-written descriptions and each entity may be referenced in multiple ways. For further comparisons of RTFM and \taskname, including why we do not simply extend RTFM, please see Appendix \ref{sec:comp_rtfm}.

%% file: sections/model.tex
\section{The \modelname{} Model}
\begin{figure}[t]
    \centering
    \includegraphics[width=0.85\linewidth]{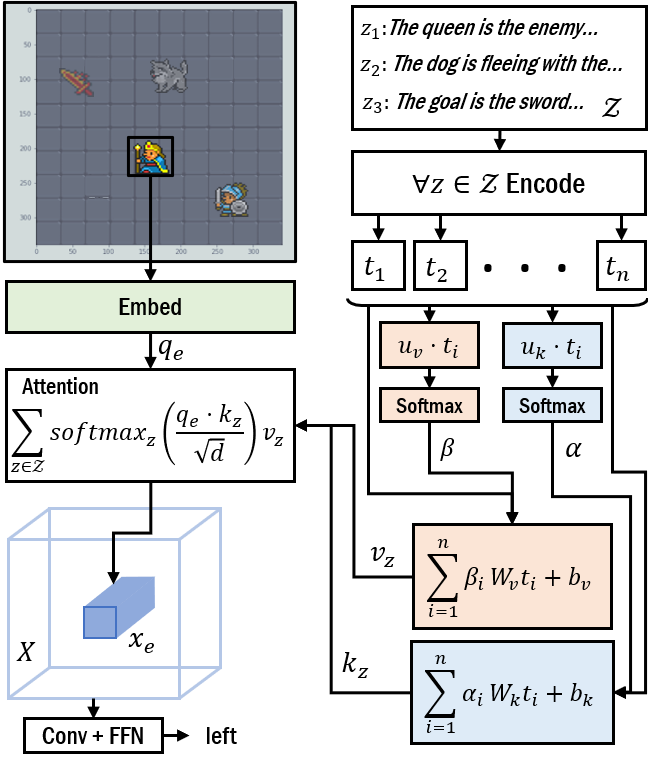}
    \caption{Schematic of our model \modelname, which creates a representation for entities using multi-modal attention over the observations and text manual. Mechanisms for the key, query, and value are shaded in blue, green, and red respectively.
    }
    \label{fig:model}
\end{figure}

As we saw in the previous section, an agent must learn to map entities to their corresponding references in the natural language manual in order to perform well in \taskname.
To learn this mapping, we develop a new model, \modelname{} (Entity Mapper with Multi-modal Attention), which employs a soft-attention mechanism over the text descriptions. At a high level, for each entity description, \modelname{} first generates key and value vectors from their respective token embeddings obtained using a pretrained language model. Each entity attends to the descriptors via a symbol embedding that acts as the attention query. Then, instead of representing each entity with its embedding, we use the resulting attention-scaled values as a proxy for the entity. This approach helps our model learn a control policy that focuses on entity roles (e.g.~\roleref{enemy}, \roleref{goal}) while using the entities' identity (e.g.~\idref{queen}, \idref{mage}) to selectively read the text. We describe each component of \modelname{} below and in Figure \ref{fig:model}.

\paragraph{Text Encoder} Our input consists of a $h\times w$ grid observation $o\in O$ with a set of entity descriptions $Z$. We encode each description $z\in Z$ using a BERT-base model whose parameters are fixed throughout training \citep{devlin2018bert, Wolf2019HuggingFacesTS}.
For a description $z$, let $t_1,...,t_n$ be its token embeddings generated by our encoder. We obtain key and value vectors $k_z, v_z$, where $\softmax$ is the softmax function:
\begin{align}
    &k_z = \sum_{i=1}^n  \alpha_iW_k t_i + b_k\label{eq:key}
    &&\alpha = \softmax\big((u_k\cdot t_j)_{j=1}^n\big)\\
    &v_z = \sum_{i=1}^n \beta_iW_vt_i + b_v\label{eq:value}
    &&\beta =\softmax\big((u_v\cdot t_j)_{j=1}^n\big)
\end{align}
The key and value vectors are simply linear combinations of $W_kt_i + b_k$ and $W_vt_i + b_v$ with weights $\alpha, \beta$ respectively, where $W_k, W_v$ are matrices which transform each token to $d$ dimensions and $b_k, b_v$ are biases. The weights $\alpha, \beta$ are obtained by taking the softmax over the dot products $(u_k\cdot t_j)_{j=1}^n$ and $(u_v\cdot t_j)_{j=1}^n$ respectively. These weights imbue our model with the ability to focus on relevant tokens. All of $W_k, b_k, u_k, W_v, b_v, u_v$ are learned parameters.

\paragraph{Entity Representation Generator}
To get a representation for each entity $e$, we embed its symbol into a query vector $q_e$ of dimension $d$ to attend to the descriptions $z\in Z$ with their respective key and value vectors $k_z, v_z$. We use scaled dot-product attention \citep{vaswani2017attention} and denote the resulting representation for the entity $e$ as $x_e$:
\begin{align}
    &x_e = \sum_{i=1}^m \gamma_i v_{z_i}\label{eq:attention}
    &\gamma = \softmax\bigg(\big(\dfrac{q_e\cdot k_{z_j}}{\sqrt{d}}\big)_{j=1}^m\bigg)
\end{align}
where $m=|Z|$ is the number of descriptions in the manual. This mechanism allows \modelname to accomplish two forms of language grounding: the key and query select relevant descriptions for each object by matching entities to names (e.g.~\strref{mage}), and the value extracts information relevant to the entities' behaviors in the world (e.g.~\roleref{enemy}, \roleref{chasing}).

For each entity $e$ in the observation, we place its representation $x_e$ into a tensor \revision{$X\in\mathbb{R}^{h\times w\times d}$} at the same coordinates as the entity position in the observation $o$ to maintain full spatial information.
The representation for the agent is simply a learned embedding of dimension $d$.

\paragraph{Action Module} To provide temporal information that assists with grounding movement dynamics, we concatenate the outputs of the representation generator from the three most recent observations to obtain a tensor \revision{$X'\in\mathbb{R}^{h\times w\times 3d}$}. To get a distribution over the actions $\pi(a | o,Z)$, we run a 2D convolution on $X'$ over the $h,w$ dimensions. The flattened feature maps are passed through a fully-connected FFN terminating in a softmax over the possible actions.
\begin{equation}
    \begin{aligned}
        y &= \mathrm{Flatten}\big(\mathrm{Conv2D}(X')\big)\\
        \pi(a | o,Z) &= \softmax \big(\mathrm{FFN}(y)\big)
    \end{aligned}
\end{equation}

In contrast to previous approaches that use global observation features to read the manual \citep{zhong2020rtfm}, we build a text-conditioned representation for each entity ($x_e$). One advantage is that $x_e$ can directly replace the entity embeddings typically used to embed the state observation in most models while still being completely end-to-end differentiable.

\camrdy{While designed for grid environments, our approach can be extended to more complex visual inputs by using CNN features as queries to extract relevant textual information for image regions, for example.} By design, \modelname \camrdy{can also learn to attend to relevant descriptions even if they reference multiple other entities. Our current version of} \taskname{} \camrdy{however, does not test for these challenges and we leave grounding entities across multiple descriptions with rich visual features to future work.} Further details about \modelname and its design can be found in Appendix \ref{sec:model_design}.

%% file: figures/train_curve.tex
\begin{figure*}[t]
    \centering
    \includegraphics[width=\linewidth]{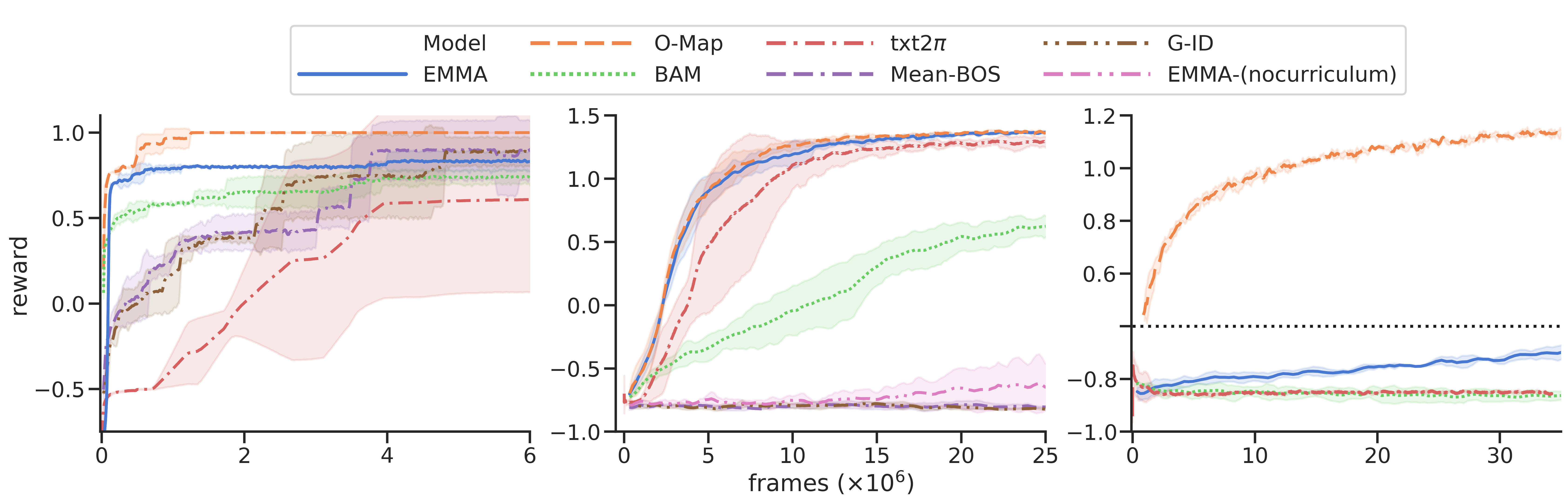}
    \caption{Average episodic rewards on S1 \textbf{(left)} S2 \textbf{(middle)} and S3 \textbf{(right)} on training games, as a function of training frames (x-axis). Note the discontinuous y-axis on S3. Reward is a combination of both single and multi-combination games. \modelname-(no curriculum) denotes \modelname trained directly on S2. Since the \bos and \gid baselines were not able to fit to S2 at all, we do not transfer them to S3. All results are averaged over three seeds and shaded area indicates standard deviation. \modelname learns faster than other baselines and on S1 and S2 almost matches the performance of \oracle. }\label{fig:training_winrates}
\end{figure*}

%% file: sections/setup.tex
\section{Experimental Setup}

\subsection{Baselines}
\paragraph{1) Mean-Bag of Sentences (Mean-BOS)} This is a variant of \modelname with the attention mechanism ablated. We average the value vectors obtained from equation \ref{eq:value} for each descriptor to obtain $\bar v$ which is used by the action module.
\begin{equation}\label{eq:avg_value}
\begin{aligned}
    \bar v = \frac{1}{|Z|}\sum_{z\in Z} v_z, \quad
    y &= \mathrm{Flatten}\big(\mathrm{Conv2D}(\mathrm{Emb}(o))\big)\\
    \pi(a | o, Z) &= \mathrm{softmax} \big(\mathrm{FFN}([y; \bar v])\big)
\end{aligned}
\end{equation}

\paragraph{2) Game ID-Conditioned (G-ID)}
To assess the importance of language in our setup, we test a model with no language understanding on \taskname. We provide an auxillary vector $I$ where each dimension corresponds to a role. $I$ is then populated with the entity symbols that reveal the mapping between entities and roles (Fig. \ref{fig:gid}, Appendix).
These symbols are embedded and concatenated to form the vector $v_I$ which is used by the action module to generate a distribution over the next actions.
\begin{equation}
\begin{aligned}
    y &= \mathrm{Flatten}\big(\mathrm{Conv2D}(\mathrm{Emb}(o))\big)\\
    \pi(a | o, Z) &= \mathrm{softmax}\big(\mathrm{FFN}([y; v_I])\big)\label{eq:gid}
\end{aligned}
\end{equation}

\paragraph{3) Bayesian Attention Module (BAM)} To assess the extent that co-occurrence statistics can help models learn $M$, we train a naive Bayes classifier to learn $M$. This approach is similar to word alignment models used in machine translation such as the IBM Model 1~\citep{brown1993mathematics}. Specifically, for some set of observed entities $E'\subseteq E$ in the current environment:
\begin{equation}
\begin{aligned}
    \mathrm{BAM}(z, E') &= \argmax_{e\in E'}P(e|z)\\
    P(z|e) = \prod_{t\in z}P(t|e), \quad
    P(t|e) &= \frac{C(t,e)}{\sum_{t'} C(t',e)}
\end{aligned}
\end{equation}
where $t\in z$ are tokens in $z$, $t'$ is any token in the manual vocabulary and $C$ refers to co-occurence counts.
We let $x_e = v_z$ from equation \ref{eq:value} for the $z$ that maps to $e$. \camrdy{By construction, $M$ is random for BAM on SC games. Note that other models can still learn $M$ using environment rewards on SC games.}
 
\paragraph{4) Oracle-Map (\oracle)}
To get an \emph{upper-bound} on performance, we consider a model that has access to the descriptor to entity map $M$, similar to \citet{narasimhan2018grounding}. This is identical to \modelname except that the representation for each entity $x_e$ is obtained as in equation \ref{eq:oracle}.
\begin{equation}
    x_e = \sum_{z\in Z} \mathds{1}[M(z)=e] v_z\label{eq:oracle}
\end{equation}

\paragraph{5) \txtpi} This method was introduced by \citet{zhong2020rtfm} alongside RTFM and features successive layers of bidirecional feature-wise modulation ($\mathrm{FILM}^2$) to model multi-hop reasoning.
Unlike RTFM, \taskname has only one text (the manual), hence we replace \txtpi's inter-text attention with self-attention.
Moreover, \txtpi does not have explicit state-tracking because it is able to identify the next correct action based on the current observation in RTFM.
This is not possible in \taskname, hence we add a state-tracker LSTM to \txtpi before the first $\mathrm{FILM}^2$ layer.
Unlike other baselines that embed each fact independently, \txtpi does not explicitly distinguish between facts.
Instead, it ingests the manual as a concatenated string of facts.

\subsection{Curriculum}
We introduce three stages of \taskname with progressive difficulty. On all stages, we train our models in a multi-task fashion by sampling a random game and appropriate manual at the start of each episode.
\paragraph{Stage 1 (S1)}  There are three entities corresponding to the \roleref{enemy}, \roleref{message} and \roleref{goal} with three corresponding descriptions. All entities begin two steps from the agent and are immovable. The agent either begins with or without the message and must interact with the correct entity. It is provided a reward of $1$ if it does so, and $-1$ otherwise.
\paragraph{Stage 2 (S2)} The same set of entities as stage 1 are present in stage 2, but entities are mobile and the agent always begins without the message. In each training game there is one chasing, one fleeing and one immovable entity. \camrdy{On test there may be any combination of movement types to force agents to adapt to unseen transition distributions} $P(s_{t+1}|s_t,a_t)$.
\paragraph{Stage 3 (S3)} In this stage there are 5 entities total with 6 descriptions, featuring one extraneous descriptor. On top of the \roleref{enemy}, \roleref{message} and \roleref{goal} entities present in stages 1 and 2, there are two additional copies of the \roleref{message} and \roleref{goal} entities, which are \roleref{enemies} and must be disambiguated by their different dynamics (e.g.~\strref{the chasing mage is an enemy} and \strref{the fleeing mage is the goal.}).

\camrdy{Human performance computed from expert playthroughs on S1, S2, and S3 are 98\%, 98\%, and 84\% respectively (see Appendix} \ref{sec:envrion_details} for details). 
Learning the entity groundings directly on stage 2 or 3 of \taskname proved to be too difficult for the models we consider.
Thus, we introduce a three-stage curriculum to train our models \citep{bengio2009curriculum}. Additional details regarding the training setup can be found in Appendix \ref{sec:imp_details}.



%% file: figures/train_table.tex
\begin{table*}
\centering
\caption{Win rates ($\pm$ stddev.) over three seeds on train. \textit{All}, \textit{MC}, and \textit{SC} denote overall, multi and single-combination games respectively. The performance of random agent subject to the same step limit on S1, S2, S3 is 7.8\%, 2.1\% and 1.6\% respectively.}
\label{tab:train_winrates}
\vskip 0.1in
\resizebox{\linewidth}{!}{
\begin{small}
\begin{sc}
\begin{tabular}{cccccccccc} 
\toprule
       & S1-All          & S1-MC           & S1-SC           & S2-All          & S2-MC           & S2-SC           & S3-All            & S3-MC             & S3-SC             \\ 
\midrule
\gid    & $89\pm 3.8$     & $90\pm 5.5$     & $89\pm 3.7$     & $3.6\pm 0.6$    & $3.4\pm 0.7$    & $3.9\pm 1.5$    & $-$               & $-$               & $-$               \\
\bos    & $90\pm 7.2$     & $91\pm 6.5$     & $90\pm 6.8$     & $2.1\pm 0.5$    & $2.9\pm 1.4$    & $2.4\pm 0.6$    & $-$               & $-$               & $-$               \\
\bayes  & $84\pm 1.3$     & $97\pm 0.9$     & $51\pm 1.6$     & $69\pm 1.1$     & $85\pm 0.9$     & $22\pm 4.8$     & $1.4\pm 0.3$      & $1.6\pm 0.5$      & $1.6\pm0.8$       \\
\txtpi  & $\bm{98\pm2.1}$ & $\bm{98\pm2.9}$ & $\bm{99\pm1.7}$ & $94\pm3.5$      & $95\pm2.1$      & $94\pm4.0$      & $3.0\pm0.6$       & $2.9\pm0.5$       & $2.8\pm0.3$       \\
\modelname   & $88\pm 2.3$     & $88\pm 2.4$     & $87\pm 1.6$     & $\bm{95\pm0.4}$ & $\bm{96\pm0.2}$ & $\bm{95\pm0.5}$ & $\bm{22\pm 3.8}$ & $\bm{21\pm 3.6}$ & $\bm{19\pm2.9}$  \\ 
\hdashline
\oracle & $97\pm 0.8$     & $97\pm 0.3$     & $96\pm0.6$      & $96\pm 0.8$     & $96\pm 0.4$     & $94\pm 0.4$     & $85\pm 1.5$       & $86\pm 0.7$       & $85\pm0.7$        \\
\bottomrule
\end{tabular}
\end{sc}
\end{small}
}
\end{table*}

%% file: sections/results.tex
\section{Results}

\subsection{Multi-Task Performance}\label{sec:multitask_perf}
Figure \ref{fig:training_winrates} shows rewards on training games as a function of training frames. The advantage of textual understanding is clear; on both S1 and S2, \modelname and \oracle converge to good policies much faster than the other baselines. However, all models except \oracle were not able to fit to S3. While \modelname can map the correct subset of descriptions to each entity, it struggles to disambiguate the descriptions based on movement dynamics. Doing so requires the challenge of mapping movement descriptions to observations of entity positions relative to the agent's own through multiple frames. Furthermore, \modelname cannot fit onto S2 without pretraining on S1 (Fig. \ref{fig:training_winrates}) due to longer episode lengths. These challenges demonstrate the need for further work on grounding text (1) to movement dynamics and (2) with long trajectories and sparse rewards.

Table \ref{tab:train_winrates} details win rates on the training games, with a breakdown over single (SC) and multi combination (MC) games. All models were able to fit to S1, but on S2 and S3, some models exhibited win rates close to random.
We observe that on MC games, the naive Bayes classifier can achieve competitive win rates by assigning over $99\%$ of training descriptors correctly. However, on SC games which require interactive entity grounding, win rates are up to $60\%$ lower.
\camrdy{This result highlights the importance of distinguishing entity groundings induced from co-occurrence statistics, and those learned from environment interactions.}

Our model (\modelname) can consistently win on both MC and SC games in S1 and S2, demonstrating \modelname's ability to ground entities without co-occurrences statistics between entity and text symbols to guide its grounding.
While \txtpi is able to fit to the S1 training games, it requires an order of magnitude more steps to do so compared to \modelname.
This is likely because \txtpi must learn to distinguish between facts observed as a concatenated string, while lacking an explicit entity-manual grounding module.

\subsection{Generalization}
\input{figures/test_table}




\paragraph{Test Games}
Results on test games are presented in Table \ref{tab:test_results}. The \gid, \bos and \txtpi baselines fail to generalize in all cases. Although the models have complete access to distinguishing information necessary to succeed, they overfit to entity-role assignments observed during training. \bayes demonstrates some ability to generalize to test games, but performance on games with single-combination entities are considerably lower, bringing the average down.

In contrast, \modelname wins $85\%$ of test games on S1 and S2, almost matching the performance of the \oracle model. By extracting information from the relevant descriptor for each entity, \modelname is able to considerably simplify each task — it simply needs to learn a policy for how to interact with \roleref{enemy}, \roleref{messenger} and \roleref{goal} archetypes instead of memorizing a policy for each combination of entities. This abstraction facilitates knowledge sharing between games, and generalization to unseen games. However, test performance on S3 for all models except \oracle does not exceed $10\%$.

\begin{figure}
    \centering
    \includegraphics[width=0.9\linewidth]{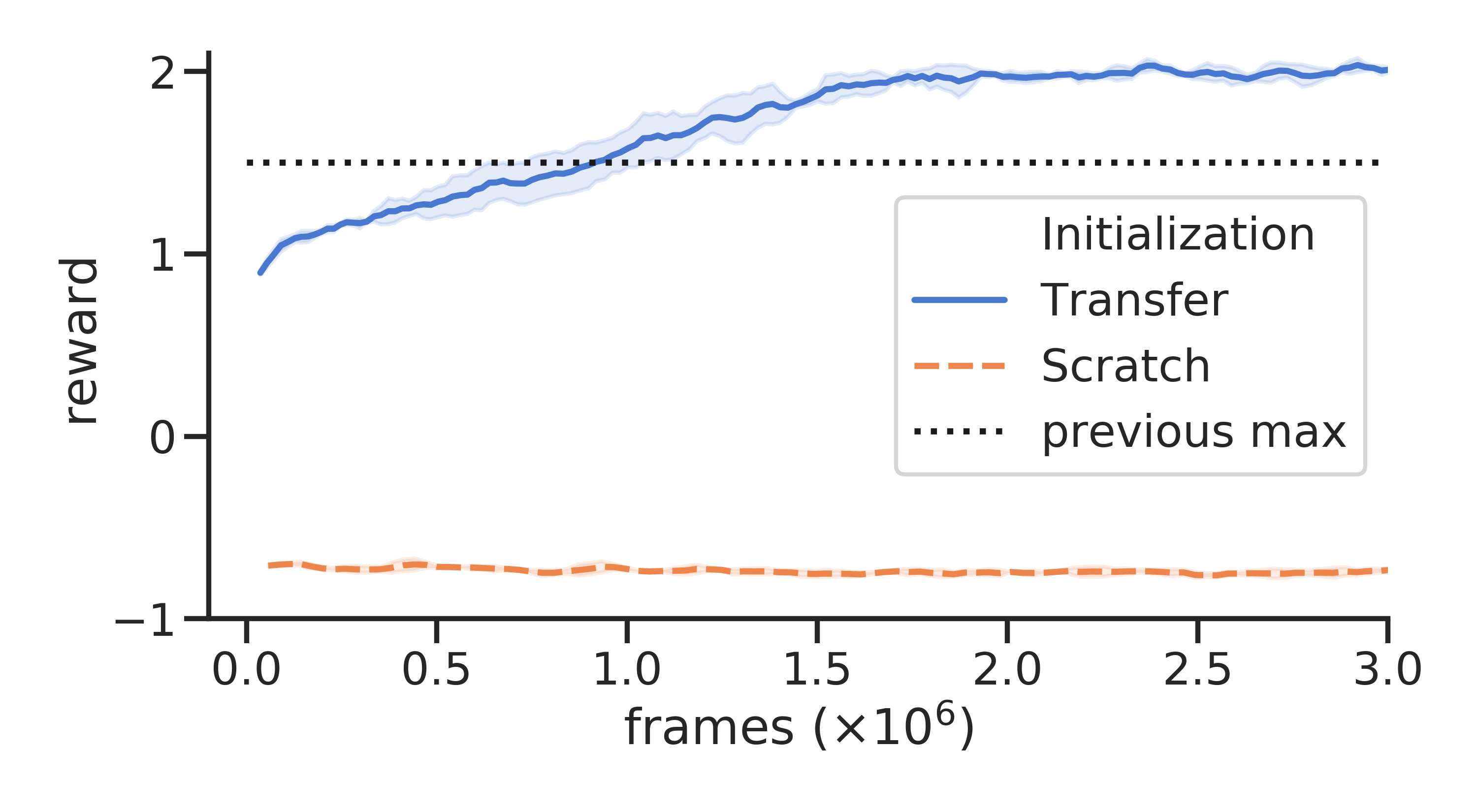}
    \caption{Transfer performance of \modelname on S2 games with novel entities and reward mechanics not found in \taskname. \modelname trained on \taskname (transfer) learns the new games much faster than a model trained from scratch (scratch).
    }\label{fig:transfer}
\end{figure}

\paragraph{New Entities} To assess \modelname's ability to pick up novel game mechanics not specified in the manual, we introduce two new stationary collectibles into \taskname --- a trap and gold which provide additional rewards of $-1$ and $1$ respectively. \camrdy{An optimal agent in this new scenario will obtain the message and also collect the gold before reaching the goal, while avoiding the enemy and the trap.} We transfer \modelname trained up to S2 onto 32 unseen games with these new entities. \modelname learns the new dynamics while accomplishing the original objectives in \taskname{} (Figure~\ref{fig:transfer}). Compared to training from scratch, \modelname pretrained on S2 achieves a higher reward in this modified setting in the same number of steps, exceeding the previous maximum reward in S2 in $1\times10^6$ steps.

\subsection{Robustness}\label{sec:robustness}
\paragraph{Train-Time}
We test \modelname's ability to learn entity groundings with added neutral entities and negated descriptions on S2 (Table \ref{tab:neg_neutral}). Due to poor performance of all models on S3, we conduct these studies on S1 and S2 only.

\input{figures/neg_neu_table}

\textit{Neutral entities}. At the start of each episode, we randomly select one of five neutral entities and insert it into the observation. The neutral entities are not described by the text, do not interact with the agent and provide no reward signal. The neutral entities are distinct from the entities in figure \ref{fig:entities}.

\textit{Negation}. On each training episode with probability $0.25$ we select one description, negate it, and change the role. (e.g.~\strref{the mage is an enemy} becomes \strref{the mage is not the message}). This case forces the model to consider the roles of the other two entities to deduce the role of the entity with the negated description. While \modelname can ground entities and performs well with neutral entities, it sometimes fails to ground the entities correctly with negated descriptions, affecting its performance on test games.

\paragraph{Test-Time} We assess the robustness of trained \bayes and \modelname models against text variations on S2 test games in table \ref{tab:app_del}. We test each model's ability to: (1) handle an extra descriptor for an entity not found in the game (Append), (2) reason about the role of objects without a descriptor by deleting a sentence from the input at random (Delete) and (3) generalize to unseen synonyms (Synonyms). For the last case, we use (unseen) templated descriptions filled in with entity synonyms not seen during training.

\input{figures/app_del_table}

Both models can retain their performance when presented with an extraneous description and suffer considerably when a description is deleted. However, \modelname generalize to unseen entity synonyms winning $75\%$ of games compared to $8.5\%$ by the \bayes model in this setting.

\subsection{Analysis of Grounding}
\begin{figure}
    \centering
    \includegraphics[width=0.9\linewidth]{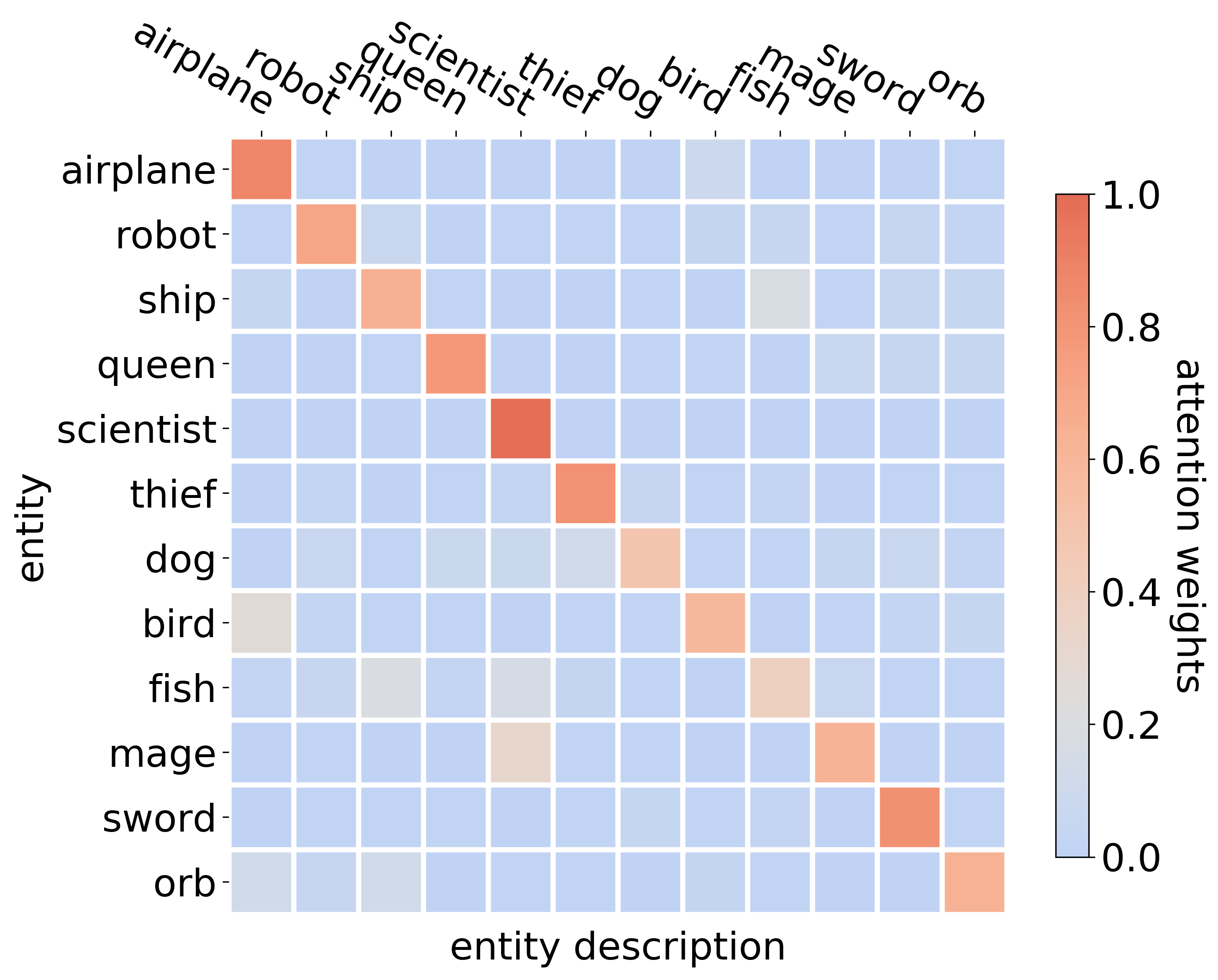}
    \caption{Attention weights for \modelname computed from equation \ref{eq:attention}. Each row is the attention weights for one entity over 12 random descriptors (one for each entity indicated by the column label). \modelname learns to map each description to the entity it references.}\label{fig:atten_heatmap}
\end{figure}
We visualize the attention weights for \modelname in Figure \ref{fig:atten_heatmap}. To assess the overall latent mapping learned by our model, we evaluate the attention weights over 12 descriptions, one for every entity. \modelname places most weight for entity $e$ onto its descriptor $z_e$. In particular, \modelname learns a grounding for \idref{dog}, \idref{bird}, \idref{fish}, \idref{mage}, \idref{sword} and \idref{orb} --- entities for which co-occurrence statistics provide no meaningful alignment information, demonstrating that our model can learn groundings for these entities via interaction alone.


%% file: figures/test_table.tex
\begin{table}
\centering
\caption{Win rates ($\pm$ stddev.) on test games over three seeds. \modelname achieves win rates on S1 and S2 test games competitive with \oracle, but performance on S3 is significantly lower.}
\label{tab:test_results}
\vskip 0.1in
\fontsize{9pt}{9pt}
\begin{sc}
\begin{tabular}{cccc} 
\toprule
       & S1-Test          & S2-Test          & S3-Test            \\ 
\midrule
\gid    & $18\pm 8.2$      & $5.2\pm 0.2$      & $-$                \\
\bos    & $6.7\pm 2.8$     & $4.7\pm 0.5$      & $-$                \\
\bayes  & $66\pm 1.5$      & $41\pm 1.7$      & $2.7\pm 0.9$       \\
\txtpi  & $2.5\pm1.7$      & $0.3\pm0.08$      & $2.6\pm0.3$        \\
\modelname   & $\bm{85\pm 1.4}$ & $\bm{85\pm 0.6}$ & $\bm{10\pm 0.8}$  \\ 
\hdashline
\oracle & $97\pm 0.3$      & $87\pm 1.8$      & $80\pm 1.5$        \\
\bottomrule
\end{tabular}
\end{sc}
\end{table}

%% file: figures/neg_neu_table.tex
\begin{table}
    \caption{Win rates ($\pm$ stddev.) over three seeds for \modelname on negation (Neg) and neutral (Neu) cases on S1 and S2.}
    \vskip 0.1in
    \centering
    \fontsize{9pt}{9pt}
    \begin{sc}
    \begin{tabular}{c c c c c}
        \toprule
        & S1-Neu & S2-Neu & S1-Neg & S2-Neg\\
        \midrule
        Train & $92\pm 1.0$ & $95\pm 0.4$ & $87\pm 3.8$ & $88\pm 8.8$\\
        Test & $88\pm 0.7$ & $78\pm 2.5$ & $67\pm 29$ & $59\pm 33$\\
        \bottomrule
    \end{tabular}
    \end{sc}
    \label{tab:neg_neutral}
\end{table}

%% file: figures/app_del_table.tex
\begin{table}
    \centering
    \caption{Win rates ($\pm$ stddev.) on S2 test games over three seeds for Append, Delete and Synonym cases for \modelname and \bayes.}
    \vskip 0.1in
    \fontsize{9pt}{9pt}
    \begin{sc}
    \begin{tabular}{c c c c}
        \toprule
        & Append & Delete & Synonyms\\
        \midrule
        \bayes & $36\pm 1.7$ & $17\pm 2.0$ & $8.5\pm 1.0$\\
        \modelname & $\bm{78\pm 3.9}$ & $\bm{33\pm 2.3}$ & $\bm{75\pm 3.4}$\\
        \bottomrule
    \end{tabular}
    \end{sc}
    \label{tab:app_del}
\end{table}

%% file: sections/conclusion.tex
\section{Conclusion}
In this paper, we introduce a new environment \taskname which does not provide prior knowledge connecting text and state observations --- the control policy must simultaneous learn to ground a natural language manual to symbols and dynamics in the environment.
We develop a new model, \modelname (Entity Mapper with Multi-modal Attention) that leverages text descriptions for generalization of control policies to new environments. \modelname employs a multi-modal entity-conditioned attention module and learns a latent grounding of entities and dynamics using only environment rewards. Our empirical results on \taskname demonstrate that \modelname shows strong generalization performance and robust grounding of entities. However, the hardest stage of \taskname which requires grounding language to subtle differences in movement patterns remains difficult for \modelname and other state of the art models. We hope our work will lead to further research on generalization for RL using natural language. 


\section*{Acknowledgements}
We are grateful to Ameet Deshpande, Jens Tuyls, Michael Hu, Shunyu Yao, Tsung-Yen Yang, Willie Chang and anonymous reviewers for their helpful comments and suggestions. We would also like to thank the anonymous AMT workers for their indispensable contributions to this work. This work was financially supported by the Princeton SEAS Senior Thesis Fund.

%% file: sections/appendix.tex
\section{Text Manual}
\label{sec:text_details}
\begin{table}[ht]
    \caption{Example template descriptions. Each underlined word in the example input indicate blanks that may be swapped in the template. Each template takes a word for the object being described (bird, thief, mage), its role (enemy, message, goal) and an adjective (dangerous, secret, crucial).}
    \label{tab:example_template}
    \vskip 0.1in
    \small
    \centering
    \resizebox{\linewidth}{!}{%
    \begin{tabular}{p{\linewidth}}
        \textbf{Example Input}\\
        \hline
        - The \underline{bird} that is coming near you is the \underline{dangerous} \underline{enemy}. \\
        - The \underline{secret} \underline{message} is in the \underline{thief}'s hand as he evades you. \\
        - The immovable object is the \underline{mage} who holds a \underline{goal} that is \underline{crucial}.\\
        \hline
        \smallbreak
        \textbf{Enemy Descriptions}\\
        \hline
        Adjectives: dangerous, deadly, lethal\\
        Role: enemy, opponent, adversary\\
        \hline
        \smallbreak
        \textbf{Message Descriptions}\\ 
        \hline
        Adjectives: restricted, classified, secret\\
        Role: message, memo, report\\
        \hline
        \smallbreak
        \textbf{Goal Descriptions}\\
        \hline
        Adjectives: crucial, vital, essential\\
        Role: goal, target, aim\\
        \hline
    \end{tabular}%
    }
\end{table}
To collect the text manual, we first crowdsource 82 templates (with 2,214 possible descriptions after filling in the blanks). Each Amazon Mechanical Turk worker is asked to paraphrase a prompt sentence while preserving words in boldface (which become the blanks in our templates). We have three blanks per template, one each for the entity, role and an adjective. For each role (enemy, message, goal) we have three role words and three adjectives that are synonymous (Table \ref{tab:example_template}). Each entity is also described in three synonymous ways. Thus, every entity-role assignment can be described in 27 different ways on the same template. Raw templates are filtered for duplicates, converted to lowercase, and corrected for typos to prevent confusion on downstream collection tasks.

\begin{table}
    \centering
    \caption{Example descriptions for \taskname{} after the second round of data collection. Note the use of synonyms \textit{flying machine} and \textit{airplane}, which also needs to be disambiguated from \textit{winged creature} (bird). Some descriptions have information divided across two separate sentences. We do not correct typos (\textit{italics}). Some typos (\textit{plan} instead of \textit{plane}) render the description useless, forcing the agent to infer the correct entity form the other descriptions in the text manual.}\label{tab:example_turk}
    \vskip 0.1in
    \small
    \resizebox{\linewidth}{!}{%
    \begin{tabular}{p{\linewidth}}
        \hline
        - the flying machine remains still, and is also the note of upmost secrecy. \\
        - the airplane is coming in your direction. that airplane is the \textit{pivitol} target. \\
        - the winged creature escaping from you is the vital target.\\
        - the fleeing \textit{plan} is a critical target.\\
        \hline
    \end{tabular}%
    }
\end{table}

To collect the free form text for a specific entity-role assignment, we first sample a random template and fill each blank with one of the three possible synonyms. The filled template becomes the prompt that is shown to the worker. For each prompt, we obtain two distinct paraphrased sentences to promote response diversity.

On all tasks (template and free-form) we provide an example prompt (which is distinct from the one provided) and example responses to provide additional task clarity.  Aside from lower-casing the free-form descriptions and removing duplicate responses, we do no further preprocessing.

To ensure fluency in all responses, we limited workers to those located in the United States with at least 10,000 completed HITs and an acceptance rate of $\geq99\%$. Some representative responses of free-form responses are presented in table \ref{tab:example_turk}. We paid our workers US\$0.25 for each pair of sentences, as we found the task was usually finished in $\leq$ 1 min. This translates to a pay of at least \$15 per hour per worker.

\section{Environment Details}
\label{sec:envrion_details}
\begin{table}[ht]
    \caption{Basic information about our environment \taskname{}. Each game features 3 out of 12 possible unique non-agent entities, with up to 5 non-agent entities total. Each entity is assigned a role of enemy, message or goal.
    }
    \label{tab:task_stats}
    \vskip 0.1in
    \centering
    \resizebox{\linewidth}{!}{%
        \begin{tabular}{p{0.3\linewidth}|p{0.7\linewidth}}
        \textbf{Entities} & bird, dog, fish, scientist, queen, thief, airplane, robot, ship, mage, sword, orb \\
        \hline
        \textbf{Roles} & enemy, message, goal\\
        \hline
        \textbf{Movements} & chasing, fleeing, immovable \\
        \hline
        \textbf{Total games} & $P(12,3) = 1320$\\
    \end{tabular}%
    }
\end{table}

Details about \taskname can be found in table \ref{tab:task_stats}. On stage 1 (S1), the three entities start randomly in three out of four possible locations, two cells away from the agent. The agent always begins in the center of the grid. It starts without the message with probability $0.8$ and begin with the message otherwise. When the agent obtains the message, we capture this information by changing the agent symbol in the observation.

On stage 2 (S2), the agent and entities are shuffled between four possible starting locations at the start of each episode. On S2, the mobile entities (fleeing, chasing) move at half the speed of the agent.  On S2 train, there is always one chasing, one fleeing and one immovable entity. Test games can feature any combination of movement dynamics.

On stage 3 (S3), the agent and non-player entities are shuffled between 6 possible starting locations. As with S2, entities move at half the speed of the agent. The one distractor description may either reference the enemy as a \roleref{message} or a \roleref{goal}, with a movement type that is distinct from the true movement type of the enemy. S3 test games do not feature unseen movement combinations, since the movements of the entities are integral to the gameplay in S3.

Since there are only 4 single-combination (SC) training games and 40 multi-combination (MC) training games, we sample the games non-uniformly at the start of each episode to ensure that there is enough interaction with SC entities to induce an entity grounding. On all stages we sample an SC game with probability $0.25$ and an MC game otherwise. Not all descriptions have movement type information (e.g.~\strref{chasing}). We also collect \textit{unknown type} descriptions with no movement type information. During training, in S1 and S2, each description is independently an unknown type description with probability $0.15$. On S3, we do not provide any description with no movement information, since this would render disambiguation via movement differences impossible.

\begin{figure}[t]
    \centering
    \includegraphics[width=\linewidth]{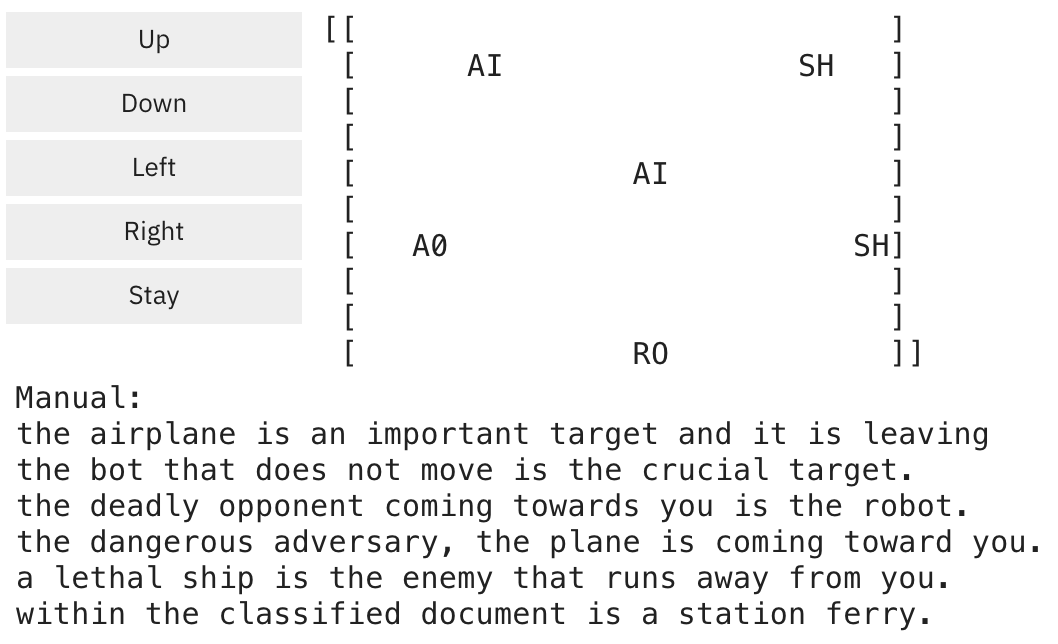}
    \caption{An S3 game on the interface used to collect human playthroughs. \idref{A0} represents the agent and other entities are represented by the first two letters of the entity name in Table \ref{tab:task_stats}.}\label{fig:human_play}
\end{figure}

\paragraph{Human Playthroughs} We collect expert human playthroughs using the interface presented in Figure \ref{fig:human_play}. The human expert has access to the manual, navigation commands, and a text-rendered grid observation. The grid observation uses the first two letters of the entity name from Table \ref{tab:task_stats} to represent each entity. Thus, human performance does not reflect the challenge of grounding entities by playing the environment; rather it quantifies the difficulty of completing the task with entity groundings provided upfront.

\begin{figure}[t]
    \centering
    \includegraphics[width=0.85\linewidth]{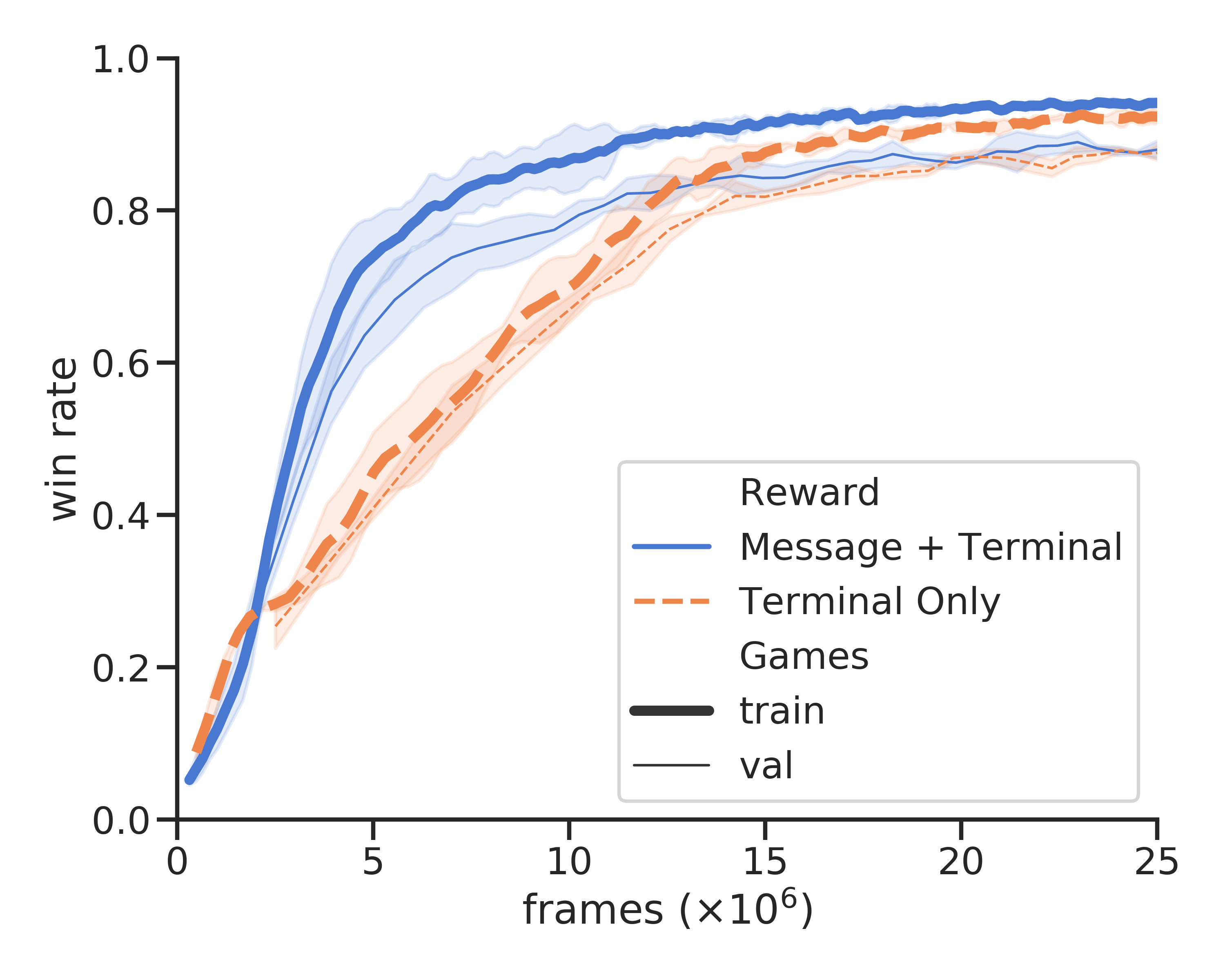}
    \caption{Win rates of \modelname on \taskname with intermediate rewards (Message + Terminal) and terminal rewards only (Terminal Only) on S2 games. Results are over three seeds and shaded area indicates standard deviation. }\label{fig:terminal_rewards}
\end{figure}

\paragraph{Terminal Rewards} On S2, we provide an intermediate scalar reward of $0.5$ for obtaining the message. To assess whether only terminal rewards is sufficient for \modelname to learn a good policy on \taskname, we evaluate \modelname on S2 using $\pm1$ terminal rewards in Figure \ref{fig:terminal_rewards}. Intermediate rewards help \modelname converge to a higher win rate slightly faster, but \modelname can converge to the same win rate using just terminal rewards.

\paragraph{Negation} We procedurally generate the negated text by negating existential words (e.g.~\strref{is an enemy} becomes \strref{is not an enemy}). We manually negate those descriptions not captured by the rules. During both training and evaluation, we provide a complete text manual without any negated description with $0.75$ probability, and randomly select a description in the manual to negate otherwise. When we negate an entity description $z_e$ to $z_e'$, we also change the role (\strref{...is an enemy} becomes \strref{...is not a goal}, for example). Thus the information present in the manual has not changed, but the agent must look at the remaining two descriptions to deduce the role of $e$ with description $z_e'$.

\paragraph{Transfer Learning} We test transfer by introducing two new entities -- a trap and a gold which provide rewards of $-1$ and $1$ respectively. Both collectables are randomly shuffled between two possible starting locations at the start of each episode and do not move. We train the models in this new setting in a multi-task fashion on the 32 validation games. After the agent encounters either the trap or gold, the collected item disappears. Neither item terminates the episode and the agent can still win or lose the current episode regardless of whether it has picked up the gold or trap.

\subsection{Comparison with RTFM}\label{sec:comp_rtfm}
The main novelty of our work (both the \taskname environment and our model) is in specifically tackling the issue of \textit{entity grounding} without any prior knowledge. To do this, \taskname in contrast to RTFM (1) does not have any signal connecting entities to text references, (2) features much richer language, and (3) requires interaction in the environment to ground entities to text. We describe these in more detail:
\begin{enumerate}
    \item RTFM’s observation space consists of a grid of text in which entity names are identical to their corresponding references in the manual. Thus, both the text in the manual and the observation are embedded into the same space (e.g.~using the same word vectors), essentially providing models with the entity grounding upfront. In contrast, our environment has a separate set of symbols for the entities with no relation to the text in our manual. Thus, the entities and text are embedded into different spaces, and learning to map between these two spaces is the key challenge in our environment that has not been explored before.
    \item RTFM features only 32 total rule-based templates for the text, and each entity can only be referred to in a single way (\idref{goblin} is always \strref{goblin}). In contrast, we crowdsourced thousands of completely free-form descriptions in \textit{two} rounds using Amazon Mechanical Turk. After obtaining the seed templates from the first round, we intentionally inject multiple synonyms for each entity to construct each prompt for the second round. Workers often further paraphrased these synonyms, resulting in 5, 6 or often more ways to describe the same entity (e.g.~\strref{airplane}, \strref{jet}, \strref{flying machine}, \strref{aircraft}, \strref{airliner} all describe \idref{plane}.). The need to map these different text references to the same entity symbol further complicates the entity grounding problem in our case and more closely mirrors the challenges of grounding entities in the real world. We believe \taskname provides a much closer approximation to natural language compared to RTFM.
    \item RTFM features all possible combinations of entities during training which provides an additional signal that may simplify the grounding problem.
    \item Each entity in RTFM only moves in a single way, whereas in \taskname, each entity may have different dynamics such as fleeing, chasing, and immovable entities (and this is also described in the text). This also allows us to test our model’s ability to generalize to unseen dynamics with unseen entity movement combinations, whereas in RTFM the evaluation on unseen games is essentially state-estimation.
\end{enumerate}

\taskname shares many aspects with RTFM (e.g.~grid-world with different entities and goals). That said, there are numerous reasons why we were not able to adapt the original RTFM environment to meet our requirements. We enumerate them here:
\begin{enumerate}
    \item The dynamics in RTFM make entity grounding (the primary focus of our work) difficult. \taskname requires much simpler reasoning than RTFM, and it is already too difficult to ground entities directly in \taskname without a curriculum. RTFM sidesteps the issue by providing this grounding beforehand.
    \item Obtaining enough crowdsourced descriptions is hard with RTFM because of the more complicated dynamics. In RTFM, there are monsters, weapons, elements, modifiers, teams, variable goals and different weaknesses between entity types that need to be specified. Collecting enough descriptions that are entirely human written would be challenging. (RTFM sidesteps this issue by using templates to generate their text manual).  In contrast, there are only entities, 3 roles, and a fixed goal in \taskname, making the text-collection task much more tractable.
    \item The entities in our \taskname environment are carefully chosen to make entity grounding harder. In RTFM, each entity is referred to in a single way, and it is not clear how to refer to them in multiple ways (e.g.~there are not too many other ways to say \strref{goblin}). In contrast, we specifically chose a set of entities that allowed for multiple ways of description, and actively encouraged this during data collection.
    \item The combination of entities that appear during training in \taskname is carefully designed. This is so that we can introduce single-combination games and the associated grounding challenges that come with it.
    \item We have different movement types for each entity. These different movements are referred to in our text manual and significantly increase the richness and variety of descriptions we collected, and also allow us to test generalization to unseen movement combinations. In RTFM, the entity movements are the same and fixed for all entities.
    \item Each entity’s attribute is referenced in the observation in RTFM, e.g.~the grid has entries such as \idref{fire goblin}. We could add to the cell an extra symbol for \idref{fire}, but this further obfuscates the entity grounding problem we are focusing on, because we would also need to obtain a grounding for all the attributes such as \idref{fire}.
\end{enumerate}
\color{black}

\section{Implementation and Training Details}
\label{sec:imp_details}
All models are end-to-end differentiable and we train them using proximal policy optimization (PPO) \citep{schulman2017ppo} and the Adam optimizer \citep{kingma2014adam} with a constant learning rate of $5\times 10^{-5}$. We also evaluated learning rates of $5\times 10^{-4}$ and unroll lengths of 32 and 64 steps by testing on the validation games. On S1, S2 and S3 we limit each episode to 4, 64, and 128 steps respectively and provide a reward of $-1$ if the agent does not complete the objective within this limit. Note that the computation of random agent performance is also subject to these step constraints.

\begin{figure}[h]
    \centering
    \includegraphics[width=\linewidth]{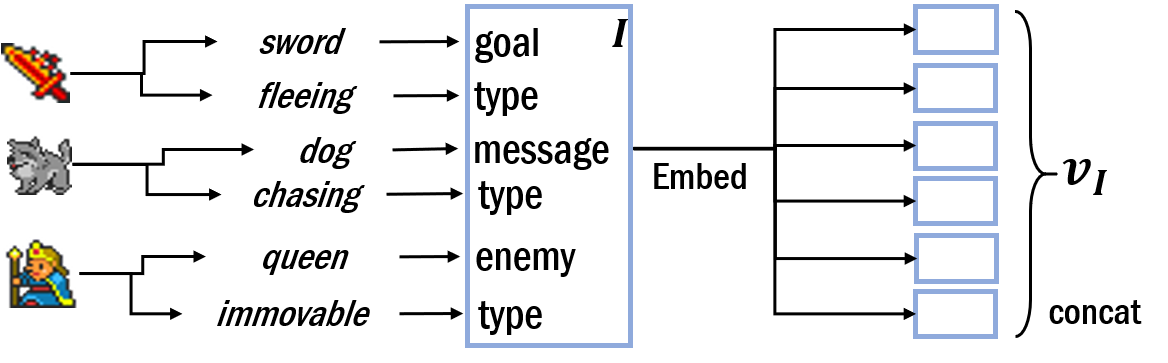}
    \caption{G-ID model}\label{fig:gid}
\end{figure}

For all experiments we use $d=256$. When multiple entities $E'$ overlap in the observation, we fill the overlapping cell with the average of the entity representations $\frac{1}{|E'|}\sum_{e\in E'}x_{e}$. The convolutional layer consists of $2\times 2$ kernels with stride $1$ and $64$ feature maps. The FFN in the action module is fully-connected with 3 layers and width of $128$. To give the \bos and \gid baselines (Fig. \ref{fig:gid}) the ability to handle the additional conditioning information, we introduce an additional layer of width $512$ at the front of the FFN for those baselines only. Between each layer, we use leaky ReLU as the activation function.

We pretrain BAM on $1.5\times 10^6$ episodes. If two descriptions map to the same entity, we take the one with higher $P(e|z)$, and If an entity receives no assignment we represent it with a learned default embedding $\mathrm{Emb}(e)$. \txtpi is trained using 10-12 actors, a model dimension of 128, and a learning rate of 0.0002.

We train models for up to 12 hours on S1, 48 hours on S2 and 72 hours on each S3. We use the validation games to save the model parameters with the highest validation win rate during training and use these parameters to evaluate the models on the test games. All experiments were conducted on a single Nvidia RTX2080 GPU.

\section{Model Design}\label{sec:model_design}

The weights $u_k$ and $u_v$ were introduced to make sure that the token embeddings for filler words such as \strref{the}, \strref{and}, \strref{or} do not drown out the words relevant to the task when we take the average in equations \ref{eq:key} and \ref{eq:value}. Qualitatively, we observe that $u_k$ learns to focus on tokens informative for identifying the entity (e.g.~\idref{mage}, \idref{sword}) while $u_v$ learns to focus on tokens that help identify the entities’ roles (e.g.~\idref{enemy}, \idref{message}).

We also found that using a pretrained language model was critical for success due to the large number of ways to refer to a single entity (e.g.~\strref{airplane}, \strref{jet}, \strref{flying machine}, \strref{aircraft}, \strref{airliner} all refer to \idref{plane}).

\subsection{Model Variations}
\begin{figure}[t]
    \centering
    \includegraphics[width=0.85\linewidth]{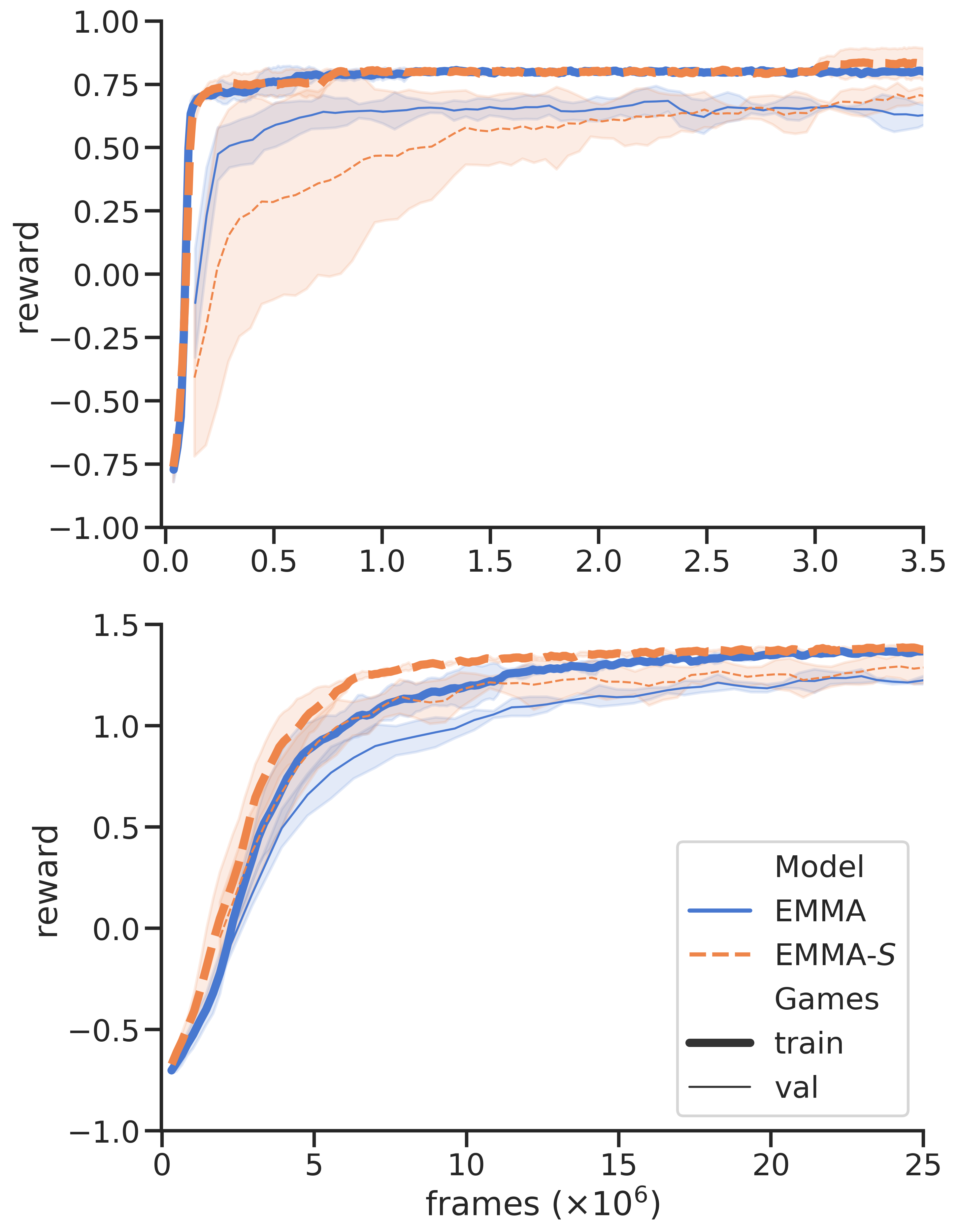}
    \caption{Average episodic rewards on S1 \textbf{(top)} and S2 \textbf{(bottom)} on training (thick line) and validation (thin line) games, as a function of training steps (x-axis) for both \modelname (solid line) and \modelname-$S$ (dotted line). Both models are able to perform well, however, \modelname is able to obtain a good validation reward faster. All results are averaged over three seeds and shaded area indicates standard deviation. }\label{fig:training_sigma}
\end{figure}

\begin{figure}[t]
    \centering
    \includegraphics[width=0.85\linewidth]{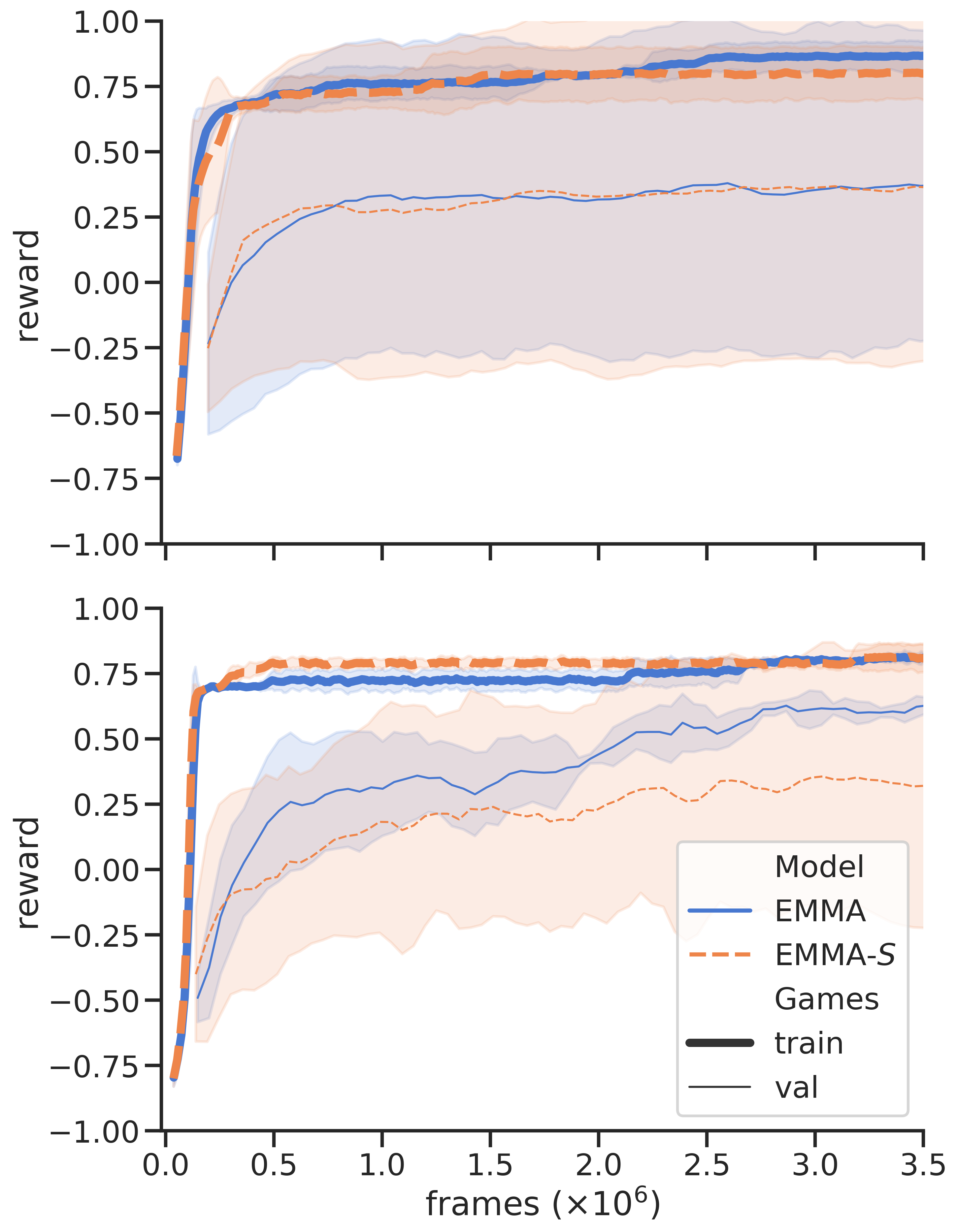}
    \caption{Average episodic rewards on S1 games with negation \textbf{(top)} and neutral entities \textbf{(bottom)} on training (thick line) and validation (thin line) games, as a function of training steps (x-axis) for both \modelname (solid line) and \modelname-$S$ (dotted line). Both models struggle on negation, but \modelname is able to perform well with neutral entities. All results are averaged over three seeds and shaded area indicates standard deviation. Note the shared x-axis.}\label{fig:neu_neg_sigma}
\end{figure}

We consider a variation to \modelname. Instead of obtaining token weights $\alpha, \beta$ in equations \ref{eq:key} and \ref{eq:value} by taking a softmax over the token-embedding and vector products $u_k\cdot t$ and $u_v\cdot t$, we consider independently scaling each token using a sigmoid function. Specifically, we will obtain key and value vectors $k_z$ and $v_z$ using:
\begin{align}
    &k_z = \sum_{i=1}^n \frac{S(u_k\cdot t_i)}{\sum_{i=1}^nS(u_k\cdot t_i)}W_k t_i + b_k \label{eq:key_sig}\\
    &v_z = \sum_{i=1}^n \frac{S(u_v\cdot t_i)}{\sum_{i=1}^n S(u_v\cdot t_i)}W_vt_i + b_v\label{eq:value_sig}
\end{align}
where $S$ is the logistic sigmoid function, and all other details are identical to \modelname. We call this model \modelname-$S$. We notice that both \modelname and \modelname-$S$ are able to obtain good training and validation performance, whith \modelname-$S$ obtaining  higher rewards on S2. However, on S1, \modelname is able to obtain a higher validation reward faster (Fig. \ref{fig:training_sigma}). Moreover, \modelname can learn robust groundings even with neutral entities, while \modelname-$S$ often overfits to a spurious grounding with neutral entities (Fig. \ref{fig:neu_neg_sigma}). Although the independent scaling in \modelname-$S$ allows the model to consider more tokens simultaneously, the softmax selection of \modelname facilitates more focused selection of relevant tokens, and this may help prevent overfitting.

\subsection{Comparison with Transformer}
\begin{figure}[t]
    \centering
    \includegraphics[width=0.85\linewidth]{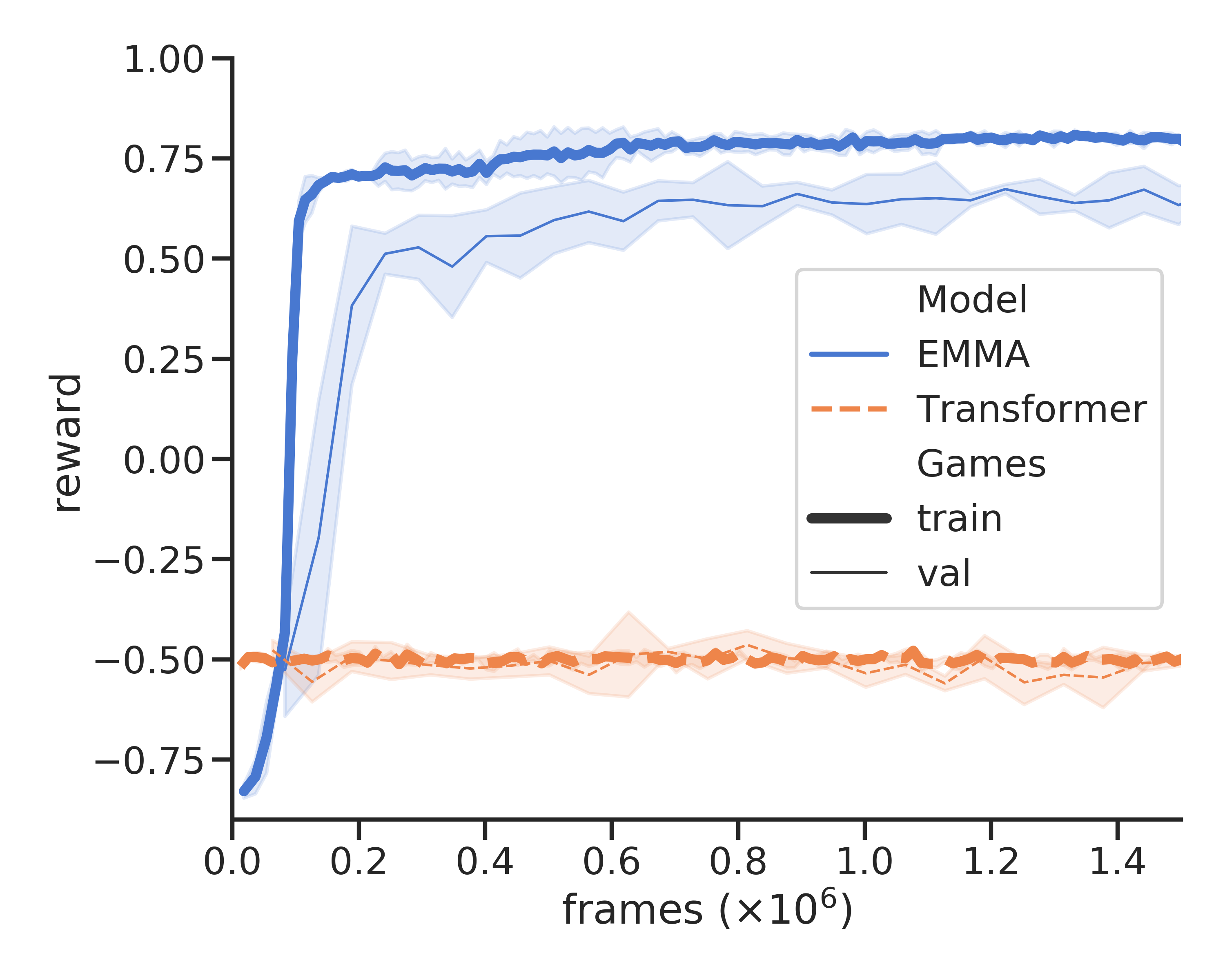}
    \caption{Average episodic rewards on S1 games with as a function of training steps (x-axis) for both \modelname (solid line) and a baseline agent consisting of a BERT model that ingests the manual and state observation converted to a string (dotted line). While \modelname is able to fit to both training and validation games, the transformer baseline struggles to learn. All results are averaged over three seeds and shaded area indicates standard deviation.}\label{fig:transformer_baseline}
\end{figure}

\modelname relies heavily on the dot-product attention mechanism to extract relevant information from the text manual. To assess the extent that attention alone is sufficient for solving \taskname, we train a Transformer \citep{vaswani2017attention} on \taskname.

Specifically, we use a pretrained BERT-base model \citep{devlin2018bert} that is identical to the one used by \modelname. We first concatenate the text descriptions $d_1,...,d_n$ to form the manual string $s_m$. For each entity in the observation, we generate a string $s_e$ by indicating the $x$ and $y$ coordinates for every entity $e$ as follows: \strref{$e$: $x$, $y$;}. We then convert the entire grid observation into a string $s_o$ by concatenating $s_e$ for every entity $e$ in the observation. The final input to BERT is then $s_m$ {\tt[SEP]} $s_o$. We train action and value MLPs on top of the {\tt[CLS]} representation in the final layer of the BERT model. The MLPs are identical to the ones used in \modelname. The entire model is end-to-end differentiable and we train it using PPO using an identical setup to the one used to train \modelname.

The results of training this Transformer baseline on S1 is presented in Figure \ref{fig:transformer_baseline}. While \modelname is able to fit to both training and validation games, the rewards for the Transformer baseline do not significantly increase even after $1.5\times10^6$ steps. We hypothesize that the difficulty of encoding spatial information in text form makes it very difficult for this model to learn a performant policy on \taskname.